\documentclass[final,5p,times,twocolumn,numbers]{elsarticle}
\usepackage{caption}
\usepackage{amsmath} 
\usepackage[bookmarks=false]{hyperref}
    \hypersetup{colorlinks,
      linkcolor=blue,
      citecolor=blue,
      urlcolor=blue}

\usepackage{tgpagella}
\usepackage{balance}

\usepackage{amssymb}

\usepackage[figuresright]{rotating}

\begin{document}
\begin{frontmatter}




\title{LakotaBERT: A Transformer-based Model for Low Resource Lakota Language}


\author{Kanishka Parankusham} 
\author{Rodrigue Rizk}
\author{KC Santosh\corref{cor1}}

\address{AI Research Lab, Department of Computer Science, University of South Dakota\\ 
414 E Clark St, Vermillion, SD 57069, USA}

\begin{abstract}
Lakota, a critically endangered language of the Sioux people in North America, faces significant challenges due to declining fluency among younger generations.  This paper introduces LakotaBERT, the first large language model (LLM) tailored for Lakota, aiming to support language revitalization efforts. Our research has two primary objectives: (1) to create a comprehensive Lakota language corpus and (2) to develop a customized LLM for Lakota. We compiled a diverse corpus of 105K sentences in Lakota, English, and parallel texts from various sources, such as books and websites, emphasizing the cultural significance and historical context of the Lakota language. Utilizing the RoBERTa architecture, we pre-trained our model and conducted comparative evaluations against established models such as RoBERTa, BERT, and multilingual BERT. Initial results demonstrate a masked language modeling accuracy of 51\% with a single ground truth assumption, showcasing performance comparable to that of English-based models. We also evaluated the model using additional metrics, such as precision and F1 score, to provide a comprehensive assessment of its capabilities. By integrating AI and linguistic methodologies, we aspire to enhance linguistic diversity and cultural resilience, setting a valuable precedent for leveraging technology in the revitalization of other endangered indigenous languages.
\end{abstract}

\cortext[cor1]{\textit{E-mail}: kc.santosh@usd.edu\\ \\}

\end{frontmatter}
\vspace*{-6pt}

\section{Introduction}
\label{main}

Transformer-based Large Language Models (LLMs) have been taking over the world for the past five years. Many neural machine translation (NMT) models have surpassed statistical machine translation models by a large margin \cite{Devlin_2018, Rizk_2022}. NMT systems excel at generating fluent translations because of their autoregressive nature, but they require a large amount of parallel datasets. However, Lakota is a low-resource language, with limited annotated datasets and computational resources for natural language processing (NLP) applications, and as demonstrated by Koehn and Knowles (2019), building neural models for low-resource languages can be quite challenging \cite{Koehn_2017}.

The Lakota people are Native Americans living in North and South Dakota. In 1805, the population was estimated to be 8,500; now, the population is 170,000, but sadly, there are only about 2,000 speakers of the Lakota language \cite{Lakota_Consortium_2016}. It is termed a "definitely endangered language" by UNESCO \cite{UNESCO_2024}. In recent years, revitalization efforts for the language have increased, driven by members of the Lakota community. Organizations like the Lakota Language Consortium \cite{Lakota_Consortium_2024}, a nonprofit made up of community leaders, linguists, educators, and volunteers, are at the forefront of these efforts. Their mission is to support the revitalization and preservation of Lakhótiyapi by working to increase the total number of new speakers. They focus on developing language learning materials, training teachers, and promoting a vibrant and positive language community. Since 2004, these materials have assisted over 20,000 students across more than 60 school systems in the region and have contributed to healthy language-learning communities. Their teacher training programs have expanded community capacity for advanced instruction in the language, utilizing effective methods and integrating contextual learning and grammar. Overall, their goal is to support the language-learning community with the materials, resources, and events needed to speak and utilize the Lakota language in everyday life.

\begin{figure}[tbp]
\centerline{\includegraphics[width=\columnwidth]{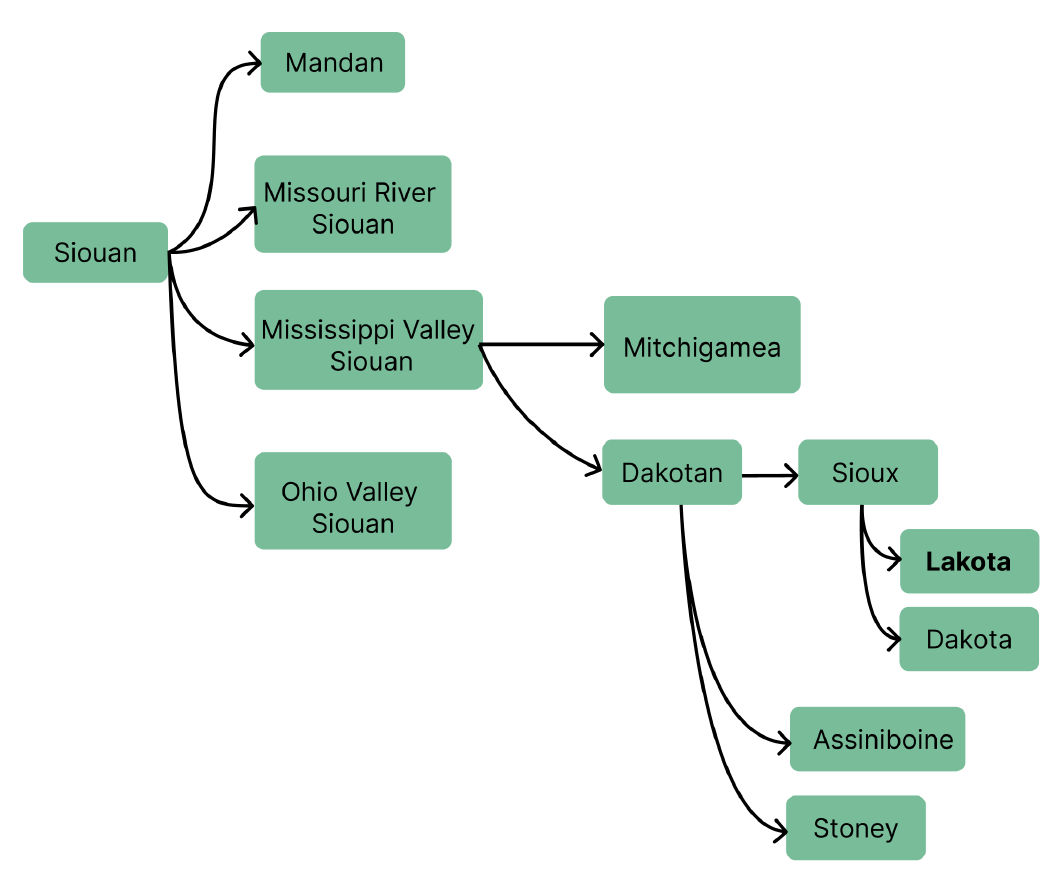}}
\caption{Siouan Language Family Tree}
\label{fig:tree}
\end{figure}

To revitalize Lakota, there have been many programs ranging from schooling to increased community engagement. However, having just a grammar book for teaching is not enough; people need resources they can interact with in various ways. This motivates us to build a large language model for Lakota to bridge this gap. In this paper, we present our efforts in collecting a Lakota dataset containing various texts, resulting in approximately 105K lines of Lakota and English, along with around 700 lines of parallel datasets. The datasets were gathered from both bilingual and monolingual sources.

The Lakota language in the dataset primarily uses Standard Lakota Orthography, a writing system developed by the Lakota Language Consortium that consistently represents each sound with a single symbol. Lakota is one of the largest Siouan languages spoken by the Sioux tribes, which are similar but distinct, as illustrated in Figure \ref{fig:tree}, which shows various Siouan languages. The Siouan-Catawban language family is a group of approximately 18 languages spoken across North America, ranging from the Appalachian Mountains in the east to the Canadian Prairies in the west. Traditionally, the Lakota language did not have a written form until the late 19th century.

One of the hallmarks of Lakota is its agglutinative morphology, meaning that words are formed by adding suffixes to roots. These suffixes can indicate a variety of grammatical functions, such as tense, aspect, mood, and negation. Lakota is also a polysynthetic language, which means that a single word can express a complete thought \cite{Ullrich_2008}. For example, the word Kichicagipi means "they avoid each other out of respect." The complexity of Lakota grammar extends to its semantics as well. Lakota verbs can encode various spatial and temporal relationships, allowing speakers to specify whether an action is taking place at a close, medium, or far distance, as well as whether it is momentary, habitual, or ongoing. As shown in Table \ref{table:lakota}\footnote{The source language and translation were scanned using Tesseract OCR, which may contain inaccuracies.} \cite{Ullrich_2008}, Lakota differs significantly from English despite sharing the same Latin script. This richness and complexity in Lakota grammar and semantics reflect the deep connection between the Lakota people and their land and way of life. The basic word order in Lakota is subject-object-verb, which can be changed for expressive purposes \cite{Van_Valin_1977}.

\begin{table}[tbp]
\centering
\small
\captionof{table}{Lakota language compared to English}
\begin{tabular}{p{0.20\linewidth}  p{0.70\linewidth}} 
 \hline
 Source & \itshape\mdseries Ehánni. hékta waníyetu óta, wičháša ithánčhan na načá ksápapi
        kin wichóh'an théča unspéunkič ičhiyapi kta chaiyounpaštakapi.\\
 Translation & \itshape\mdseries Long time ago, many years back, our wise chiefs 
             headmen encouraged us to learn new ways.\\ 
 \hline
\end{tabular}
\label{table:lakota}
\end{table}

Indigenous languages worldwide are at risk of extinction, with the Lakota language being one such example. Although Lakota holds cultural significance, its computational treatment has been sparse due to the lack of large-scale labeled datasets and resources. Many NLP tools and methods are developed primarily for high-resource languages, making them unsuitable for low-resource languages like Lakota. In this paper, we present LakotaBERT, a specialized transformer model aimed at addressing the unique challenges faced by the Lakota language. 

Our work contributes to the field of low-resource language processing by developing a robust pipeline for training a transformer-based model tailored to Lakota. We believe that the methods and insights in this paper can serve as a blueprint for future work on other underrepresented languages.

In the following sections of this paper, we delve into related work in the field of low-resource languages, focusing specifically on the challenges associated with Lakota language revitalization. We will also present the findings from our own research, in which we developed a masked language model tailored for the Lakota language. This model not only showcases the potential of LLMs in supporting language preservation efforts but also details the methodology we employed for data collection. Through this exploration, we aim to highlight both the challenges and the innovative solutions available for revitalizing low-resource languages like Lakota.

\section{Related Works}
While machine translation has made significant strides in recent years, particularly for high-resource languages, low-resource languages have often been overlooked. The advent of LLMs and transformer-based neural machine translators has opened up new possibilities for these underrepresented languages. Recently, we have witnessed the emergence of numerous large language models \cite{Antoun_2020, Muller_2022, Raffel_2020}, which utilize clean datasets and more contextual content, such as paragraphs, to enhance their understanding. However, research on low-resource languages is also gaining momentum. Models like chrEn \cite{Zhang_2022} have demonstrated the feasibility of training effective models using relatively small parallel datasets, showcasing their potential through comparisons of statistical machine translation (SMT) and NMT approaches. Furthermore, innovative applications, such as deciphering ancient texts with transformers \cite{Assael_2019, Parsons_2024, Gutherz_2023}, highlight the versatility of these models in addressing challenging linguistic tasks.

To revitalize Lakota, various programs for immersion studies and degree courses have been established. In 2011, Sitting Bull College (Fort Yates, North Dakota, Standing Rock) and the University of South Dakota began degree programs aimed at creating effective Lakota language teachers. Students can earn a Bachelor of Arts in Education at the University of South Dakota or a Bachelor of Science in Education at Sitting Bull College, majoring in ``Lakota Language Teaching and Learning" as part of the Lakota Language Education Action Program (LLEAP) \cite{University_of_South_Dakota_2012}. Additionally, in 2012, Lakota immersion classes were provided for children in an experimental program at Sitting Bull College on the Standing Rock Reservation, where children speak only Lakota for their first year \cite{Donovan_2012}. Several other universities have also started offering similar programs \cite{Gauer_2021}.

Universities have partnered with organizations such as the Lakota Language Consortium (LLC) to support the new generation of speakers in a digital way. The LLC offers various tools, including an online dictionary, Owóksape (learning place), an e-learning portal, a Lakota keyboard, and a media player with Lakota audio and pronunciations spoken by native speakers, which is crucial since Lakota is primarily an oral language. The LLC also provides various short stories in Lakota, along with many other resources available for learning the language. 
Our work extends these efforts to the Lakota language, which has not previously seen significant development of language models. In future work, we aim to leverage the power of LLMs for other downstream tasks and attract the NLP community to help save and revitalize this endangered language.

\section{Data Collection and Preprocessing}

This section introduces the Lakota-English bilingual dataset used for model training. The data was collected from diverse sources, including dictionaries and web-scraped text \cite{LakotaDictionary_2023}. 
A bilingual corpus was created, consisting of approximately 105,000 lines of Lakota text and 700 lines of parallel Lakota-English text to support future research and development.

Our training data comes from a bilingual dictionary containing Lakota and English words and sentences, along with web-scraped monolingual text \cite{Ullrich_2008}. The sentences were mostly contributed by individuals from the regions of Pine Ridge, Rosebud, Cheyenne River, Standing Rock, and Lower Brule; Dakhota people from Standing Rock, Crow Creek, and Yankton; and Dakhóta people from Sisseton, Devil's Lake, and Minnesota.

The dataset primarily comprises Lakota text, distinct from Dakota, despite sharing approximately 70\% of their vocabulary. While closely related, they are considered separate languages. The dataset contains nearly 8.8 million words combined from Lakota and English, as illustrated in Table~\ref{table:2}.

\begin{table}[tbp]
\centering
\captionof{table}{Total number of English and Lakota words used in our model.}
\vspace{1em}
\begin{tabular}{ c c c c } 
 \hline
 Language & Train & Test & Total \\
 \hline
 English & 5,415,886 &  161,089 & 5,576,975\\ 
 Lakota & 2,745,034 & 131,371 & 2,876,405\\
  \hline
\end{tabular}
\label{table:2}
\end{table}

Data acquisition for Lakota presented unique challenges due to its predominantly oral tradition until the 19th century. The limited availability of written materials, even in the digital age, necessitated creative approaches. While we successfully leveraged dictionaries as a starting point, the syntax inconsistencies between different sources highlighted the need for diverse data collection methods.

Initially, most of the data was available in PDF format, which posed challenges for text extraction. To extract the Lakota and English texts, we employed the Tesseract OCR engine \cite{Smith_2007}. This process required us to convert every page into image format before feeding it into Tesseract. To enhance extraction accuracy, we segmented the digital images into smaller components, allowing for more precise recognition.

\begin{figure}[tbp]
\centerline{\includegraphics[width=\columnwidth]{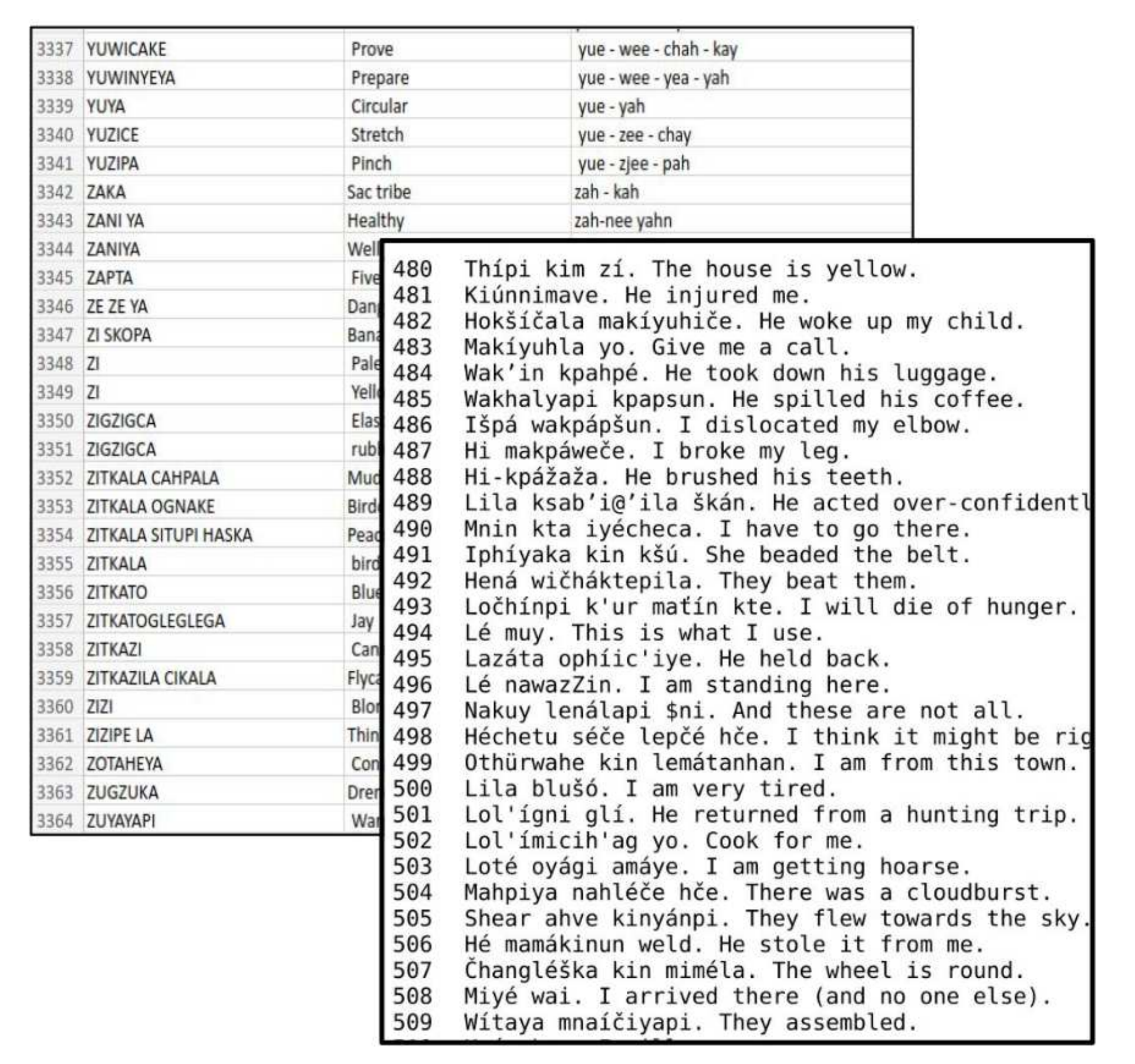}}
\caption{Collection of parallel dataset}
\label{fig:dataset}
\end{figure}

To optimize the extraction process, we initially set aside accented characters, acknowledging the importance of a subsequent refinement step to address these omissions. We also implemented a rigorous filtering system to eliminate significant amounts of unintended text before inputting it into our model. However, inconsistencies in the OCR output, as depicted in Figure \ref{fig:dataset}, continued to affect the overall quality of the dataset. This experience highlights the inherent complexities of working with OCR technology and the critical need for continuous improvements in data extraction methods.

To support our future research and development, we created a bilingual data corpus consisting of paired texts in English and Lakota. This valuable resource will enable us to train and evaluate models for various natural language processing tasks. By leveraging the corpus, we can develop more accurate and robust systems for translation, sentiment analysis, and other language-related applications.

\section{Methodology}

Given the scarcity of parallel datasets for the Lakota language, we propose a semi-supervised Robustly Optimized Bidirectional Encoder Representations from Transformers (RoBERTa) model \cite{Liu_2019} to address the challenges of processing low-resource languages. This variant of Bidirectional Encoder Representations from Transformers \cite{Devlin_2018} is built on transformer architecture \cite{Vaswani_2017} and leverages masked language modeling capabilities to effectively learn from limited data.

To create a robust training foundation for our model, we collected data from diverse sources, including bilingual dictionaries \cite{Ullrich_2008} and web-scraped texts. Our final dataset comprises approximately 105,000 lines of Lakota text and 700 lines of parallel Lakota-English text, providing a comprehensive resource for enhancing Lakota language processing.

\begin{figure*}[tbp]
\centerline{\includegraphics[width=1.9\columnwidth]{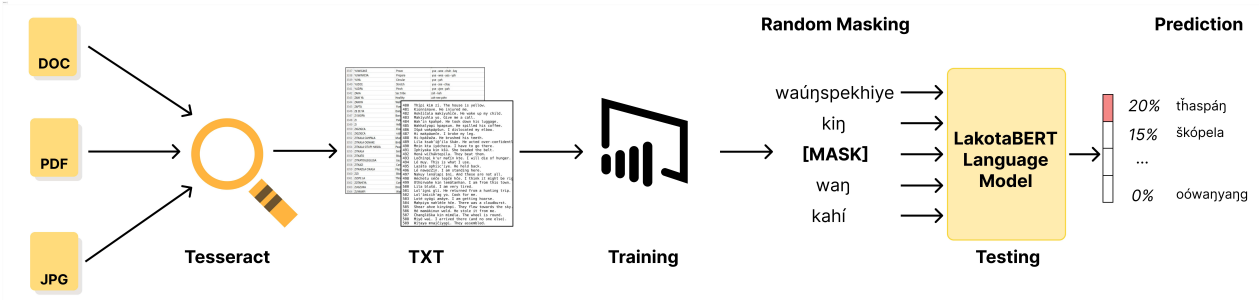}}
\caption{Overall workflow}
\label{fig:workflow}
\end{figure*}

The proposed LakotaBERT model is based on the transformer architecture, incorporating scaled dot-product attention and multi-head attention mechanisms. These features enable the model to capture complex relationships between words and phrases, allowing it to learn effectively from limited data and perform tasks such as translation and text generation. 

\begin{figure}[tbp]
\centerline{\includegraphics[width=\columnwidth]{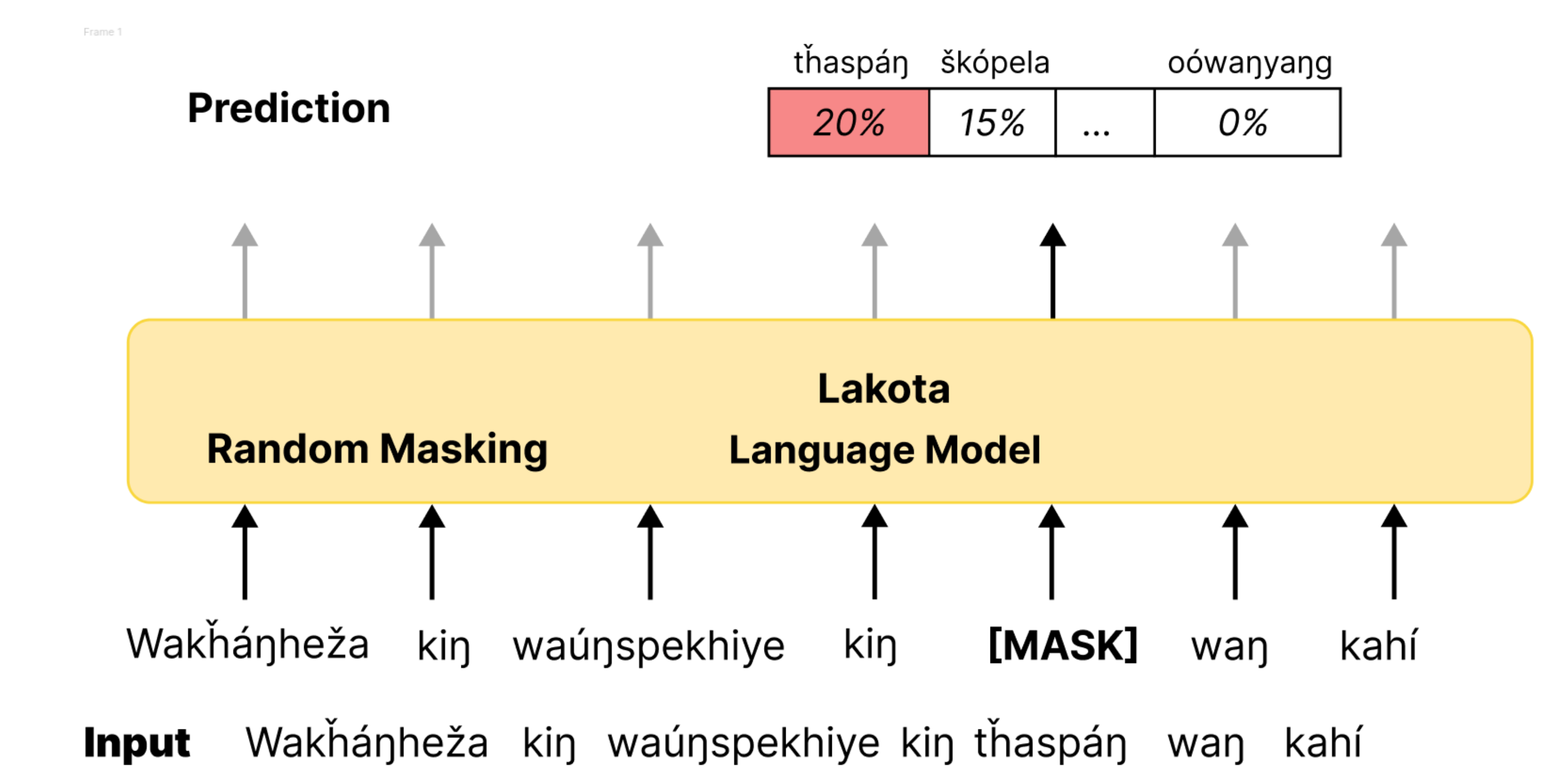}}
\caption{Input masking and prediction by the model}
\label{fig:model}
\end{figure}

Figures \ref{fig:workflow} and \ref{fig:model} outline the training and evaluation process for our semi-supervised model. Figure \ref{fig:workflow} provides an overall workflow, detailing the steps from text extraction to data preprocessing, followed by training the model and implementing masked language modeling. 
Scaled Dot-Product Attention (\ref{eq: attn}) is used for its ability to assess the relevance of different parts of an input sequence, allowing the model to focus on pertinent information and capture dependencies between words. 
\vspace{-10pt}
\begin{equation}
\mathrm{Attention}(Q, K, V) = \mathrm{softmax}\left(\frac{QK^T}{\sqrt{d_k}}\right)V,
\label{eq:attn}
\end{equation}
where \(Q\), \(K\), and \(V\) represent the query, key, and value matrices, respectively, and \(d_k\) is the dimension of the keys.

In the multi-head attention mechanism, the attention process is enhanced by allowing the model to attend to information from different representation subspaces (\ref{eq:mlt}). 
\vspace{-10pt}
\begin{equation}
\mathrm{MultiHead}(Q, K, V) = \mathrm{Concat}(\mathrm{head_1}, \ldots, \mathrm{head_h}) W^O.
\end{equation}
Each head is calculated as:
\vspace{-10pt}
\begin{equation}
\mathrm{head_i} = \mathrm{Attention}(Q W^Q_i, K W^K_i, V W^V_i),
\label{eq:mlt}
\end{equation}
where \(W^Q_i\), \(W^K_i\), and \(W^V_i\) are parameter matrices projecting the inputs into different subspaces for each attention head, and \(W^O\) combines the outputs from all heads, allowing the model to capture diverse features and relationships.

The self-attention layers in our model are connected to position-wise feed-forward networks using the GELU function to transform the output of the attention mechanism (\ref{eq:ffn}).
\begin{equation}
\mathrm{FFN}(x) = \mathrm{GELU}(xW_1 + b_1)W_2 + b_2.
\label{eq:ffn}
\end{equation}
We use embeddings to represent words as numerical vectors and employ positional encoding to convey their positions in a sequence, enabling the model to understand meanings and context for tasks such as masked language modeling.

To ensure the model learns meaningful patterns, we denoised the data before splitting it into training, validation, and testing sets. A key preprocessing step involved masking 15\% of tokens in each text with a special \(<MASK>\) token, guiding the model to learn contextual relationships and avoid superficial patterns.

For efficient representation of the Lakota text, we employed Byte Pair Encoding (BPE) during tokenization. BPE is a subword tokenization technique that merges frequently occurring character pairs, creating subword units that capture linguistic structures more effectively than traditional methods.

The model is trained on the training set, validated to optimize hyperparameters, and evaluated on the testing set to assess performance on unseen Lakota data. The masked language modeling objective plays a crucial role in learning contextual representations and improving performance. The loss function is expressed as:
\begin{equation}
L = -\frac{1}{M} \sum_{k=1}^{M} \log(p(w_k | \text{context})).
\end{equation}
This guides the training process by minimizing the discrepancy between predicted and actual word probabilities, where \(L\) represents the loss, \(M\) is the number of masked words, and \(p(w_k | \text{context})\) is the predicted probability of the masked word \(w_k\) given the context. We will evaluate our models using various metrics \cite{Manning_2008}.

The bilingual nature of our dataset, containing both Lakota and English text, offers valuable opportunities for future model development, including bilingual capabilities and translation tasks. While our current focus is on Lakota, this foundation supports future research aimed at fine-tuning translation between Lakota and English, contributing to language preservation and cultural exchange.

\section{Experiments}
\subsection{Experimental setup}
This section details the experimental setup and evaluation metrics employed to assess the performance of our semi-supervised model on the Lakota language dataset. Given the limited size of the dataset, we opted to use the TXT format for data loading. The model was trained on sequences of 512 tokens to ensure optimal performance. Training took place on an NVIDIA A100 80GB GPU \cite{NSF_1626516}, encompassing 92,000 steps, which translates to approximately 10 hours of computation. We utilized the Adam optimizer, renowned for its adaptive learning rates, incorporation of momentum, bias correction, and overall computational efficiency.


\begin{table}[tbp]
    \centering
    \caption{Hyperparameters for Lakota model}
    \vspace{1em}
    \begin{tabular}{ l c } 
        \hline
        Hyperparameters & Lakota Model \\
        \hline
        Number of parameters & 83,505,184 \\ 
        Number of layers N & 12 \\ 
        Hidden size & 768 \\
        FFN inner hidden size & 3072 \\
        Number of attention heads & 12 \\
        Attention head size & 64 \\
        Context size & 512 \\
        Vocab size & 52,000 \\
        Batch Size & 128\\
        Torch\_dtype & float32 \\
        Dropout & 0.1 \\
        Attention Dropout & 0.1 \\
        Masking Probability & 15\% \\
        \hline
    \end{tabular}
    \label{table:hyper}
\end{table}



The Lakota model employed a relatively large architecture with 12 layers, a hidden size of 768, and 12 attention heads, as detailed in Table \ref{table:hyper}. The feed-forward network, with an inner hidden size of 3072, expanded the model's representational capacity. The context size of 512 allowed the model to process sequences of up to 512 tokens, accommodating a reasonable range of sentence lengths. A vocabulary size of 52,000 was sufficient to capture the diversity of words and tokens in the Lakota language. The batch size of 128, along with dropout and attention dropout rates of 0.1, helped regularize the model and prevent overfitting. Finally, a masking probability of 15\% was used for masked language modeling, enabling the model to learn contextual representations from incomplete input sequences.

\subsection{Evaluation Metrics}

We compare LakotaBERT to other state-of-the-art baseline models, e.g. BERT, mBERT, and RoBERTa. 
We used accuracy (\ref{eq:acc}), precision (\ref{eq:prec}), MRR (\ref{eq:mrr}), character error rate (CER) (\ref{eq:cer}), hit@k\((\ k = 10\)) (\ref{eq:hit}), F1-Score (\ref{eq:f1}), and BLEU score to evaluate these models:
\vspace{-10pt}
\begin{equation}
\text{Accuracy} = \frac{\text{Correct Predictions}}{\text{Total Masked Positions}} \times 100
\label{eq:acc}
\end{equation}
\vspace{-25pt}
\begin{equation}
\text{Precision} = \frac{\text{True Positives}}{\text{True Positives} + \text{False Positives}}
\label{eq:prec}
\end{equation}
\vspace{-25pt}
\begin{equation} F1 = 2 \cdot \frac{\text{Precision} \cdot \text{Recall}}{\text{Precision} + \text{Recall}} 
\label{eq:f1}
\end{equation}
\vspace{-25pt}
\begin{equation}
\text{MRR} = \frac{1}{N} \sum_{i=1}^{N} \frac{1}{\text{Rank}_i}
\label{eq:mrr}
\end{equation}
\vspace{-25pt}
\begin{equation}
\text{CER} = \frac{1}{N} \sum_{i=1}^{N} \frac{\text{Levenshtein Distance}(true_i, pred_i)}{\max(\text{len}(true_i), \text{len}(pred_i))}
\label{eq:cer}
\end{equation}
\vspace{-25pt}
\begin{equation}
\text{Hit@K} = \frac{1}{N} \sum_{i=1}^{N} \text{I}(true_i \in \text{Top K Predictions}_i)
\label{eq:hit}
\end{equation}

\subsection{Results}
\begin{figure}[t]
    \centering
    \begin{minipage}{0.45\textwidth}
        \centering
        \includegraphics[width=\columnwidth]{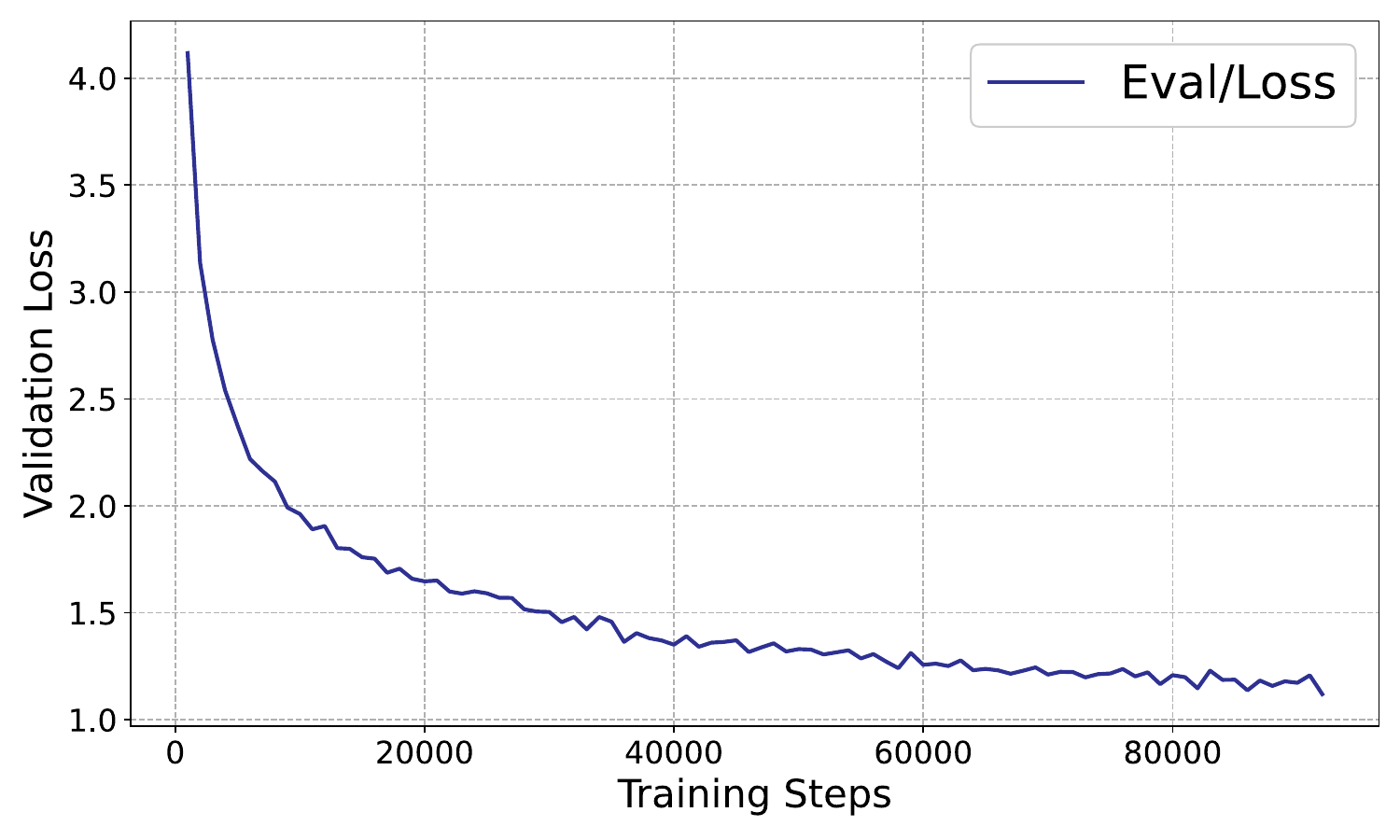}
        \caption*{(a) Tracking Evaluation Loss During Model Assessment Phases}
    \end{minipage}\hfill
    \begin{minipage}{0.45\textwidth}
        \centering
        \includegraphics[width=\columnwidth]{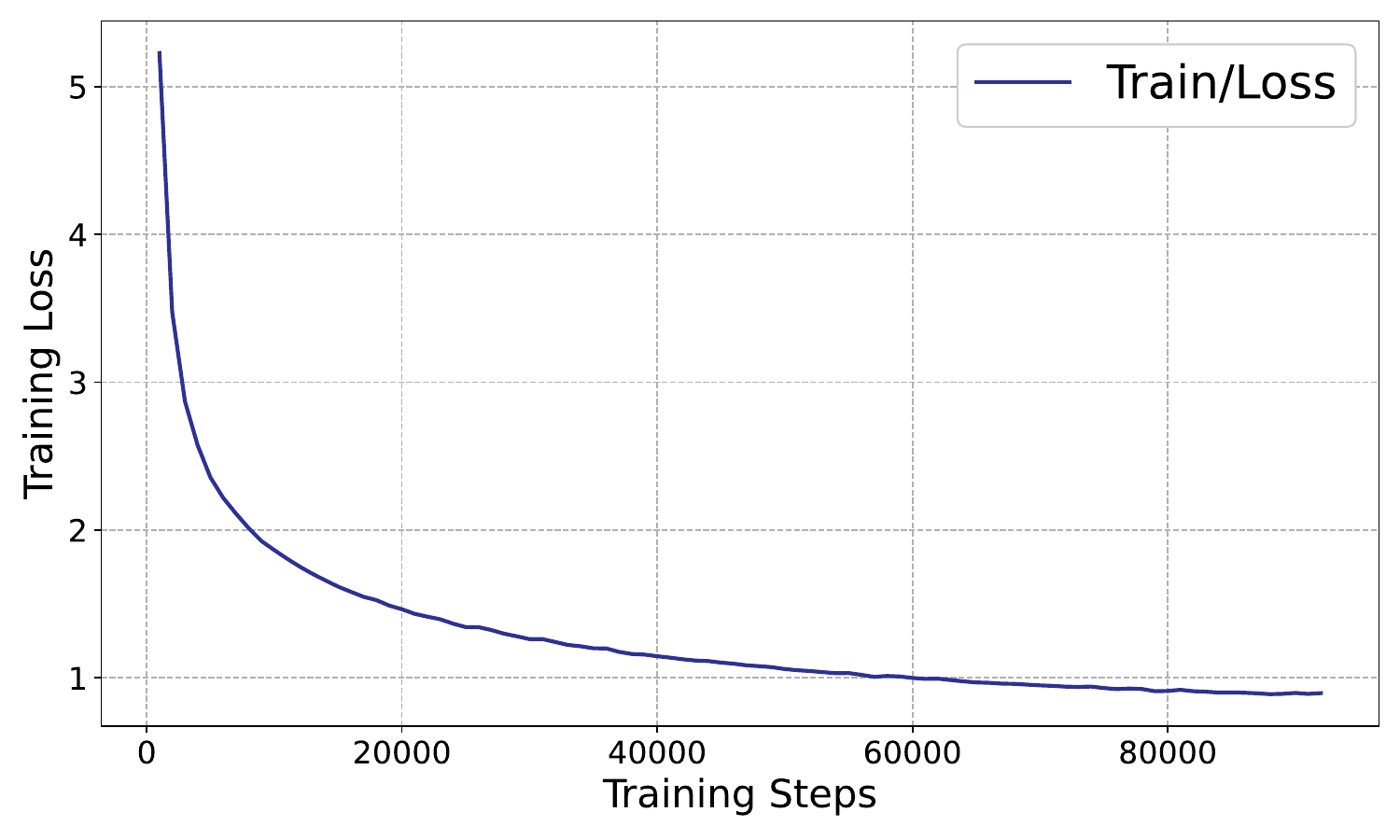}
        \caption*{(b) Tracking of Training Loss Over Epochs During Model Training}
    \end{minipage}
    \caption{Visualization of model performance across different phases, highlighting trends and patterns in the learning process.}
    \label{fig:results}
\end{figure}

The evaluation loss and training loss graphs presented in Figure \ref{fig:results} clearly illustrate the progress of our model throughout the training process. The consistent decrease in loss over time indicates that the model is successfully learning to minimize errors, reflecting its improving performance. This downward trend is a strong indicator of the model's ability to adapt and refine its predictions, suggesting that it is effectively capturing the underlying patterns in the data. 

\begin{figure}[t!]
\centerline{\includegraphics[width=\columnwidth]{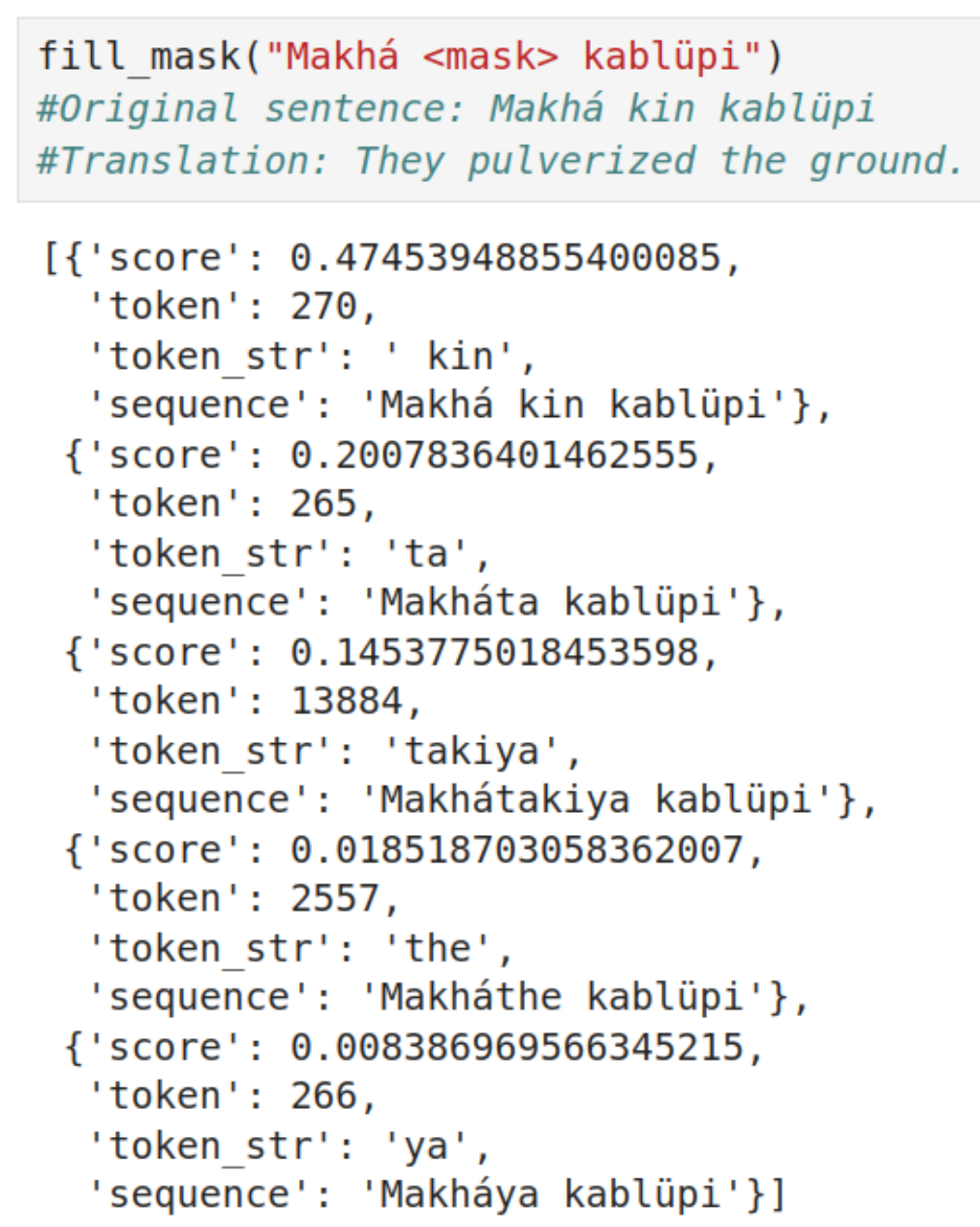}}
\caption{Perplexity scores for each generated token, along with their corresponding token numbers and the actual words produced by the LLM.}
\label{fig:masking}
\end{figure}

Upon completion of the training, Figure \ref{fig:masking} provides a detailed illustration of word masking and showcases the perplexity scores for each generated token while highlighting the actual words predicted by the language model. The perplexity scores offer valuable insights into the model's confidence in its predictions, helping to evaluate its performance in generating coherent and contextually relevant outputs. This comprehensive overview showcases the effectiveness of the language model in understanding and generating language.

Our model, LakotaBERT, achieves an accuracy of 51.48\%, slightly outperforming the baseline BERT model's accuracy \eqref{eq:acc} of 50.69\%. Although it still trails behind the more robust RoBERTa and mBERT models, which achieved accuracies of 61.15\% and 54.42\%, respectively, this performance highlights LakotaBERT's ability to capture certain nuances of the Lakota language, despite the limited available data.

In terms of precision \eqref{eq:prec}, which measures the proportion of correct predictions among all positive predictions, LakotaBERT achieved a score of 0.56. This outperformed BERT (0.47) and mBERT (0.50), although it still lags behind RoBERTa (0.61). This suggests that LakotaBERT demonstrated greater confidence in its predictions. Since there is only one ground truth per instance, precision reflects the model's accuracy in predicting that single instance, with no false positives influencing the score.

The F1-score \eqref{eq:f1}, which balances precision and recall into a single performance metric, indicates that RoBERTa performs best in this regard. Both LakotaBERT and BERT share the same F1-score, suggesting that LakotaBERT maintains a similar balance between precision and recall as BERT.

The MRR \eqref{eq:mrr}, which measures the average reciprocal ranks of the correct answers within the predicted lists, was 0.51 for LakotaBERT. This surpassed BERT's score of 0.50 and was comparable to mBERT's score of 0.54, yet still trailed behind RoBERTa (0.61). These results demonstrate LakotaBERT's ability to rank the correct answer higher among the predicted options.

The CER \eqref{eq:cer}, which measures the average character-level prediction errors normalized by the length of the longest string, was 0.43 for LakotaBERT—higher than BERT's CER of 0.42, mBERT's CER of 0.38, and RoBERTa's CER of 0.29. This indicates that LakotaBERT produced more character-level errors compared to BERT and mBERT, highlighting areas where it can improve in accurately capturing the morphological and syntactic complexities of the Lakota language.

For the Hit@K metric \((with \ K = 10\)) \eqref{eq:hit}, which evaluates the proportion of times the correct answer appears among the top K predictions, LakotaBERT achieved a score of 0.31, outperforming mBERT (0.24). This metric demonstrates LakotaBERT's ability to rank the correct answer within the top predictions, reflecting strong retrieval performance.

Finally, we observed a BLEU score of 0.09, which is commonly used in machine translation, providing an indication of the model's ability to generate translation-like outputs.

\begin{table*}[t]
    \centering
    \caption{Language Model Performance Analysis}
    \small
    \vspace{1em}
    \begin{tabular}{l c c c c c c c}
    \hline
    Model & Accuracy & Precision & F1-Score & MRR & CER & Hit@k & BLEU \\
    \hline
    BERT\cite{Devlin_2018} & 50.69\% & 0.47 & 0.50 & 0.49 & 0.42 & 0.28 & 0.09\\
    RoBERTa\cite{Liu_2019} & 61.15\% & 0.61 & 0.61 & 0.62 & 0.29 & 0.28 & 0.10 \\
    MBERT\cite{Devlin_2018} & 54.42\% & 0.50 & 0.54 & 0.51 & 0.38 & 0.24 & 0.09 \\
    LakotaBERT & 51.48\% & 0.56 & 0.49 & 0.51 & 0.43 & 0.31 & 0.09\\
    \hline
    \end{tabular}
    \label{table:1}
\end{table*}

Future research should focus on developing more tailored evaluation metrics for low-resource languages. This might involve considering factors such as linguistic diversity, cultural context, and the model's ability to generate coherent and relevant text. Additionally, exploring alternative data augmentation techniques and leveraging transfer learning from related languages can help improve model performance.

\subsection{Discussion}
LakotaBERT is the first language model built on Lakota text. Our model has demonstrated performance comparable to other models designed for the same task of masked language modeling. This indicates that the model effectively understands both Lakota and English, as it was trained on texts from both languages. The scores were calculated based on a single ground truth, which posed a challenge for all models in accurately predicting the right words. However, by considering the context of entire sentences, we can better assess the model's understanding of language. We have compiled a diverse collection of Lakota texts, creating a corpus of 105K sentences in both Lakota and English. Additionally, we have assembled a parallel dataset consisting of approximately 700 lines. This resource will be essential for future revitalization efforts of the Lakota language. With this pretrained model as a foundation, we can further fine-tune it for various tasks such as summarization, translation, and others. Human evaluation will be conducted at a later stage to ensure that the model’s outputs are culturally appropriate and resonate with native speakers, as this feedback is vital for improving the model's effectiveness and relevance. The practical implications of LakotaBERT for language revitalization are significant. By providing tools for language learning, translation, and content generation, LakotaBERT can facilitate greater access to Lakota language resources.

Our work is the first step in creating a scalable infrastructure for future NLP tasks for Lakota, which can further support language revitalization efforts. The application of LakotaBERT is not limited to academic NLP tasks but also has the potential for real-world applications, such as voice assistants, educational tools, and automated translation services, which could significantly aid in the preservation and revitalization of the Lakota language.

\section{Conclusion and Future works}
Our model, LakotaBERT, is the first of its kind to incorporate Lakota into a large language model (LLM). In addition, we have contributed a bilingual Lakota-English dataset, which includes a comprehensive corpus of 105K lines. We also implemented a Lakota model for masked language modeling. The initial evaluation results are promising, with an accuracy of 51.48\%, demonstrating that the model is successfully capturing the intricacies of Lakota's linguistic structure. This performance sets a valuable baseline for future Lakota-based models. LakotaBERT shows that even with limited annotated resources, transformer models can be effectively fine-tuned to produce competitive results on various tasks for low-resource languages like Lakota.

In future work, we plan to expand the Lakota corpus, introduce dialect-specific models, and integrate multimodal data sources, such as speech-to-text systems, to further enhance model performance. Additionally, we aim to fine-tune the model for specific language learning tasks, including translation and text generation. By leveraging both AI technology and linguistic expertise, we aspire to advance the revitalization of the Lakota language, supporting efforts to empowering future generations to connect with their linguistic heritage.

\balance

\bibliographystyle{plainnat}

\begin{thebibliography}{}




\bibitem{Devlin_2018}
Devlin, J., et al. (2018) ``Bert: Pre-training of deep bidirectional transformers for language understanding." arXiv preprint arXiv:1810.04805.

\bibitem{Rizk_2022}
R. Rizk, D. Rizk, F. Rizk, A. Kumar and M. Bayoumi, "A Resource-Saving Energy-Efficient Reconfigurable Hardware Accelerator for BERT-based Deep Neural Network Language Models using FFT Multiplication," 2022 IEEE International Symposium on Circuits and Systems (ISCAS), Austin, TX, USA, 2022, pp. 1675-1679, doi: 10.1109/ISCAS48785.2022.9937531.

\bibitem{Koehn_2017}
Koehn, P., \& Knowles, R. (2017) ``Six challenges for neural machine translation." arXiv preprint arXiv:1706.03872.

\bibitem{Lakota_Consortium_2016}
Lakota Language Consortium (LLC). (2016) ``Press Release: Lakota Language Now Critically Endangered." Lakota Language Consortium.

\bibitem{UNESCO_2024}
UNESCO. (2024) ``Lakota." UNESCO World Atlas of Languages. Retrieved from \url{https://en.wal.unesco.org/languages/lakota}

\bibitem{Lakota_Consortium_2024}
Lakota Language Consortium. (2024) ``Lakota Language Consortium - Revitalizing the Lakota Language." The Lakota Language Consortium. https://lakhota.org/

\bibitem{Ullrich_2008}
Ullrich, J. F. (2008) ``New Lakota Dictionary." Lakota Language Consortium.

\bibitem{Van_Valin_1977}
Van Valin Jr, Robert Detrick. (1977) ``Aspects of Lakota Syntax." University of California, Berkeley.

\bibitem{Stahlberg_2020}
Stahlberg, F. (2020) ``Neural machine translation: A review."  Journal of Artificial Intelligence Research, 69, 343-418.

\bibitem{Wang_2019}
Wang, Q., et al. (2019) ``Learning deep transformer models for machine translation." arXiv preprint arXiv:1906.01787.

\bibitem{Vaswani_2017}
Vaswani, A., et al. (2017) ``Attention is all you need." Advances in neural information processing systems, 30.

\bibitem{Antoun_2020}
Antoun, W., Baly, F., \& Hajj, H. (2020) ``Arabert: Transformer-based model for arabic language understanding." arXiv preprint arXiv:2003.00104.

\bibitem{Muller_2022}
Müller, M., \& Laurent, F. (2022) ``Cedille: A large autoregressive french language model." arXiv preprint arXiv:2202.03371.

\bibitem{Raffel_2020}
Raffel, C., et al. (2020) ``Exploring the limits of transfer learning with a unified text-to-text transformer." Journal of machine learning research, 21(140), 1-67.

\bibitem{Zhang_2022}
Zhang, S., Frey, B., \& Bansal, M. (2022) ``How can NLP help revitalize endangered languages? A case study and roadmap for the Cherokee language." arXiv preprint arXiv:2204.11909.

\bibitem{Assael_2019}
Assael, Y., Sommerschield, T., \& Prag, J. (2019) ``Restoring ancient text using deep learning: a case study on Greek epigraphy." arXiv preprint arXiv:1910.06262.

\bibitem{Parsons_2024}
Parsons, F. N. (2024) ``Revealing Text from a Still-rolled Herculaneum Papyrus Scroll (PHerc. Paris. 4)." Zeitschrift für Papyrologie und Epigraphik, 229, 1-13.

\bibitem{Gutherz_2023}
Gutherz, G., et al. (2023) ``Translating Akkadian to English with neural machine translation." PNAS nexus, 2(5), pgad096.

\bibitem{University_of_South_Dakota_2012}
University of South Dakota. (2012) ``Lakota Language Education Action Program."  \textit{LLEAP}. Retrieved from \url{https://www.lleap.org/USD.html}

\bibitem{Donovan_2012}
Donovan, L. (2012) ``Learning Lakota, one word at a time."  \textit{Bismarck Tribune}. Retrieved from \url{https://bismarcktribune.com/news/state-and-regional/learning-lakota-one-word-at-a-time/article_0352569a-2a97-11e2-9c5c-001a4bcf887a.html}

\bibitem{Gauer_2021}
Gauer, S. (2021) ``SDSU making strides in Lakota language revitalization efforts." The Collegian. \url{https://sdsucollegian.com/22127/lifestyles/sdsu-making-strides-in-lakota-language-revitalization-efforts/}

\bibitem{LakotaDictionary_2023} 
Lakota Dictionary. (2023) ``Lakota Dictionary" \textit{Lakota Dictionary}. Retrieved from \url{https://www.lakotadictionary.org/phpBB3/index.php}

\bibitem{Smith_2007}
Smith, R. (2007) ``An overview of the Tesseract OCR engine." Ninth international conference on document analysis and recognition (ICDAR 2007). Vol. 2. IEEE.

\bibitem{Liu_2019}
Liu, Y., et al. (2019) ``Roberta: A robustly optimized bert pretraining approach." arXiv preprint arXiv:1907.11692.



\bibitem{Sennrich_2015}
Sennrich, R. (2015) ``Neural machine translation of rare words with subword units." arXiv preprint arXiv:1508.07909.

\bibitem{Manning_2008}
Manning, C. D. (2008) ``Introduction to information retrieval."

\bibitem{NSF_1626516}
National Science Foundation. (n.d.) ``Award abstract \#1626516: MRI: Acquisition of the Lawrence Supercomputer to Advance Multidisciplinary Research in South Dakota." https://www.nsf.gov/awardsearch/showAward?AWD\_ID=1626516\&HistoricalAwards=false


\bibitem{Curtis_2014}
Curtis, C. M. (2014) ``A finite-state morphological analyzer for a Lakota precision grammar." Proc. LREC.

\bibitem{Nikitricky_2021}
Nikitricky. (2021) ``Random Paragraphs (Dataset)." Kaggle. https://www.kaggle.com/datasets/nikitricky/random-paragraphs/data

\bibitem{Wieting_2019}
Wieting, J., et al. (2019) ``Beyond BLEU: training neural machine translation with semantic similarity." arXiv preprint arXiv:1909.06694.



 \end{thebibliography}

\clearpage

\normalMode

\end{document}